\documentclass[runningheads]{llncs}
\usepackage{todonotes}

\usepackage{multirow}

\usepackage{graphicx}
\usepackage{todonotes}

\begin{document}
%
%\title{Is it primordial to use an end-to-end architecture for spoken language understanding?\thanks{This work was granted access to the HPC resources of IDRIS under the allocation 2020-AD011011838 made by GENCI.}}

\title{Where are we in semantic concept extraction for Spoken Language Understanding?
\thanks{This work was granted access to the HPC resources of IDRIS under the allocation 2020-AD011011838 made by GENCI and partially funded through the H2020 SELMA project (grant No 957017).}}

\titlerunning{Where are we in semantic concept extraction for SLU?}

%\author{First Author\inst{1}\orcidID{0000-1111-2222-3333} \and
%Second Author\inst{2,3}\orcidID{1111-2222-3333-4444} \and
%Third Author\inst{3}\orcidID{2222--3333-4444-5555}}
%

\author{Sahar Ghannay\inst{1}, Antoine Caubrière\inst{2}, Salima Mdhaffar\inst{2}, Gaëlle Laperrière\inst{2}, Bassam Jabaian\inst{2}, Yannick Estève\inst{2} }

\authorrunning{S. Ghannay et al.}

\institute{Université Paris-Saclay, CNRS, LISN, 91400, Orsay, France\\
\email{firstname.lastname@limsi.fr}
\and
LIA - Avignon Université, France \\
\email{firstname.lastname@univ-avignon.fr}\\
}
\maketitle              % typeset the header of the contribution
\begin{abstract}
Spoken language understanding (SLU) topic has seen a lot of progress these last three years, with the emergence of end-to-end neural approaches.
Spoken language understanding refers to natural language processing tasks related to semantic extraction from speech signal, like named entity recognition from speech or slot filling task in a context of human-machine dialogue.
Classically, SLU tasks were processed through a cascade approach that consists in applying, firstly, an automatic speech recognition process, followed by a natural language processing module applied to the automatic transcriptions.
These three last years, end-to-end neural approaches, based on deep neural networks, have been proposed in order to directly extract the semantics from speech signal, by using a single neural model.
More recent works on self-supervised training with unlabeled data open new perspectives in term of performance for automatic speech recognition and natural language processing.
In this paper, we present a brief overview of the recent advances on the French MEDIA benchmark dataset for SLU, with or without the use of additional data. 
We also present our last results that significantly outperform the current state-of-the-art with a Concept Error Rate (CER) of 11.2\%, instead of 13.6\% for the last state-of-the-art system presented this year.

\end{abstract}
\section{Introduction}
%\yk{Pour moi : ajouter les références bibliographiques} \\

Spoken language understanding (SLU) refers to natural language processing tasks related to semantic extraction from the speech signal~\cite{tur2011spoken}, like named entity recognition from speech, call routing, slot filling task in a context of human-machine dialogue\ldots 

Usually, SLU tasks were processed through a cascade approach that consists in applying first an automatic speech recognition (ASR) process, followed by a natural language processing module applied to the automatic transcription~\cite{de2007spoken}.
For both automatic speech recognition and natural language processing, deep neural networks (DNN) have made great advances possible, leading to impressive improvements of qualitative performance for final SLU tasks~\cite{amodei2016deep,collobert2011natural,vaswani2017attention}.

These three last years, end-to-end neural approaches, based on deep neural networks, have been proposed in order to directly extract the semantics from speech signal, by using a single neural model~\cite{serdyuk2018towards,ghannay2018end}. 
A first advantage of such approaches consists in a joint optimization of the ASR and NLP part, since the unique neural model is optimized only for the final SLU task.
Another advantage is the limitation of the error propagation: when using a cascade approach, an error in the first treatment implies errors in the following ones. In a neural end-to-end approach, the model decision is delayed to the output layer: all the information uncertainty is handled until the final decision. 

Very recently, works on self-supervised training with unlabeled data open new perspectives in term of performance for automatic speech recognition and natural language processing~\cite{baevski2020wav2vec,devlin-etal-2019-bert}. They can be applied to SLU task.

This study presents experimental results on the French MEDIA benchmark dataset.
This benchmark dataset is one of the most challenging benchmarks for SLU task.
In this paper, we present a brief overview of the performance evolution of state-of-the-art systems on this benchmark dataset.
We also present an approach that takes benefit from acoustic-based and linguistic-based models pre-trained on unlabelled data: this approach represents the next milestone to be surpassed.
%Last, we also study the results that can be reached when only the (small) SLU dataset is used, without external labelled data.

%In pipeline architecture for dialogue system, the Spoken Language Understanding system (SLU) extract a list of semantic concepts from an automatic transcription of a user speech signal.  

\section{MEDIA dataset}
\label{sec:MEDIA}
The French MEDIA corpus~\cite{BonneauMaynard2005}, is dedicated to semantic extraction from speech in a context of human-machine dialogues for a hotel booking task. 
This dataset was created as a part of the Technolangue project of the French government in 2002. Its main objective is to set up an infrastructure for the production and dissemination of language resources, the evaluation of written and oral language technologies, the participation in national and international standardisation bodies and an information monitoring in the field.

The MEDIA dataset is made of telephone dialogue recordings with their manual transcriptions and semantic annotations. 
It is composed of 1257 dialogues from 250 different speakers, collected with a Wizard-of-Oz setting between two humans: one plays a computer, the other plays the user. The dataset is split into three parts (train, dev, test) as described in table~\ref{tab:MEDIAstat}. 
In this work, we used the user part of MEDIA, since it has both speech and semantic annotations.

The semantic domain of this corpus is represented by 76 semantic concept tags such as \textit{room number, hotel name, location, etc}. Some more complex linguistic tags, like co-references, are also used in this corpus. 

The following sentence (translated from French) is an example of the MEDIA content: "I would like to book one double room in Paris up to one hundred and thirty euros". It will be annotated as (I would like to book, \textsl{reservation}), (one, \textsl{number-room}), (double room, \textsl{room-type}), (up to, \textsl{comparative-payment}), (one hundred and thirty, \textsl{amount-payment}), (euros, \textsl{currency-payment}). 

In~\cite{bechet2019benchmarking}, Béchet and Raymond showed why the MEDIA task can be considered as the most challenging SLU benchmark available, in comparison to other well-known benchmarks such as ATIS~\cite{dahl1994expanding}, SNIPS~\cite{coucke2018snips}, and  M2M~\cite{shah2018building}.

%: (1) the training set (13K sentences, 10h46 hours), (2) the development set (1.3K sentences, 1h13 hours), and (3) the test set (3.5K sentences, 2h59 hours).

% add nombre de concept ??
\begin{table}[t]
  \caption{The official MEDIA dataset distribution}
  \label{tab:MEDIAstat}
  \centering
  \begin{tabular}{| c | c | c | c | c |}
\hline    
\textbf{Data}   & \textbf{Nb}  & \textbf{Nb}  & \textbf{Nb} &  \textbf{Nb}   \\
 \textbf{} &  \textbf{Words} &  \textbf{Utterances} & \textbf{Concepts} &  \textbf{Hours}   \\\hline    
\hline    
%    train & 94274  &  13729 & 31.7k & 10h46m\\%46.70s \\
    train & 94.2k  &  13.7k & 31.7k & 10h46m\\%46.70s \\
\hline    
%   dev &  10742 & 1372  & 3.3k & 01h13m\\%1.37s\\
    dev &  10.7k & 1.3k  & 3.3k & 01h13m\\%1.37s\\
\hline    
%    test & 26614  & 3764  & 8.8 k & 02h59m\\%4.66s\\
    test & 26.6k  & 3.7k  & 8.8 k & 02h59m\\%4.66s\\
\hline    
  \end{tabular}
\end{table}

\section{Overview of approaches proposed for the MEDIA benchmark}

\subsection{Cascade approach}

Conventional SLU systems are designed as a cascade of components. Each of them solves separately a specific problem. First, an ASR module, trained on a large amount of data, maps speech signals to automatic transcriptions. This is then passed on to a natural language understanding (NLU) module that predicts semantic information from the automatic transcriptions.
In this approach, error propagation is unavoidable, despite the performance of current ASR and NLU systems. 
In addition, those modules are optimized separately under different criteria.
The ASR system is trained to minimize the word error rate (WER), while the NLU module is trained to minimize the concept error rate (CER) in case of slot filling task. This separate optimization suggests that a cascade SLU system is suboptimal.

Working on automatic transcriptions, for an SLU task on MEDIA corpus, is highly challenging. Many approaches have been proposed. 
Early NLU approaches were based on generative models such as Stochastic finite state transducers (FST), on discriminative or conditional models such as conditional random fields (CRFs) %\cite{lafferty2001conditional,raymond2007generative,hahn2010comparing} 
and support vector machines(SVMs)\cite{hahn2010comparing}. % or the joint use of generative and descriminative models \cite{dinarelli2009concept}.
In the light of the success of neural approaches in different fields, some studies developed neural architectures for SLU task. In~\cite{mesnil2013investigation}, the author presents the first recurrent neural architecture dedicated to SLU for the ATIS benchmark corpus. This neural model was applied to transcriptions despite speech signals directly.
In \cite{simonnet:hal-01526298,simonnet-etal-2018-simulating}, for the first time, an encoder-decoder neural network structure with attention mechanism~\cite{bahdanau2014neural} was proposed for this task. This time, it was on manual and automatic transcriptions from the MEDIA corpus. 
In order to reduce the unavoidable SLU performance decline due to ASR errors, the authors in \cite{simonnet:hal-01526298} have proposed ASR confidence measures to localize ASR errors. These confidence measures have been used as additional SLU features to be combined with lexical and syntactic features, useful for characterizing concept mentions. 
In \cite{simonnet-etal-2018-simulating}, the authors proposed an approach to simulate ASR errors from  manual transcriptions, to improve the performance of SLU systems. The use of the resulting corpus prepares the SLU system to ASR errors during their training and makes it more robust to ASR errors.

%\cite{yao2014spoken,Mesnil:2015:URN:2876369.2876380,vukotic:hal-01196915,dinarelli:hal-01553830,dinarelli2017effective,simonnet:hal-01433202,simonnet:hal-01526298}.

%More recently, unsupervised language representation models such as BERT~\cite{devlin-etal-2019-bert} have been shown to achieve state of the art results in different SLU tasks such as ATIS~\cite{Mandy19} and MEDIA~\cite{ghannay-2020-Coling}. 

\subsection{End-to-end approach}
\label{sec:e2e}

As seen in the previous section, one problem with cascaded approaches is the propagation of errors through the components.
The intermediate transcription is noisy due to speech recognition errors, and the NLU component has to deal with these errors. %\remsg{on dit SLU ou NLU ?? YE : NLU quand on part de la transcription, SLU sinon} %as input
The other problem comes from the separate optimizations of the different modules.

To tackle these issues, end-to-end approaches were proposed in order not to use an intermediate speech transcriptions.
This kind of approach aims to develop a single system directly optimized to extract semantic concepts from speech.

SLU end-to-end systems are usually trained to generate both recognized words and semantic tags~\cite{ghannay2018end,desot2019towards}.

Until now, mainly two kinds of neural architectures have been proposed on the MEDIA benchmark.
The first one is based on the use of the Connectionist Temporal Classification (CTC) loss function~\cite{graves2006connectionist}, while the other one is based on the use of an encoder-decoder architecture with attention mechanism~\cite{bahdanau2014neural}.

\subsubsection{CTC approach}
\label{sec:ctc_approach}
~~~\\
%\ac{Add ce qu'est la CTC loss + fonction mapping ??}
In this work, we call CTC approach the neural architecture trained by using the CTC loss function.
This loss function allows the system to learn an alignment between the input speech and the word and concept sequences to produce.

To our knowledge, the best-published results with a CTC approach on MEDIA were obtained by~\cite{Caubriere2019}.
In this study, the authors proposed a neural architecture largely inspired by the DeepSpeech~2 speech recognition system.
The neural architecture is a stack of two 2D-invariant convolutional layers (CNN), followed by five bidirectional long short term memory (bLSTM) layers with sequence-wise batch normalization, a classical fully connected layer, and the softmax output layer.
As input features, we used spectrograms of power to normalize audio clips, calculated on 20ms windows.
This system was trained following the curriculum-based transfer learning approach, which consists in training the same model through successive stages, with different tasks ranked from the most generic one to the most specific one.
The authors used speech recognition tasks, then named entity extraction and finally semantic concept extraction tasks.

\subsubsection{Encoder-decoder approach with attention mechanism}
~~~\\

The encoder-decoder architecture was initially implemented in the machine translation context.
This approach quickly showed its benefits for the speech recognition task~\cite{chorowski2014end,chan2016listen,chiu2018state}, and more recently for SLU tasks~\cite{serdyuk2018towards,9413581}.

The encoder-decoder architecture is divided into two main parts.
First, an encoder receives the speech features as input, and provides its hidden states to build a high-level representation of the features.
This high-level representation is then passed on to an attention module. It identifies the parts of these representations that are relevant for each step of the decoding process.
Next, the attention module computes a context vector from these representations to feed the decoder.
Finally, the decoder processes the input context vectors to predict the transcription of speech, enriched with semantic concepts. At each decoding time step, a new context vector is computed from the encoded speech representations.
Unlike CTC approaches, the output sequence size of an encoder-decoder approach does not depend on the input sequence size.

A recent study~\cite{9413581} used a similar architecture and obtained the state-of-the-art performance for the MEDIA task.
The encoder part is composed of four 2-dimensional convolution layers followed by four bLSTM layers.
Each convolution layer is followed by a batch normalization.
The decoder part is a stack of four bLSTM layers, two fully connected layers, and a softmax layer.
The input features of the network are 40-dimensional MelFBanks with a Hamming window of 25ms and 10ms strides.

This encoder-decoder system is trained following the curriculum-based transfer learning, with the same data used for the CTC approach presented in section~\ref{sec:ctc_approach}, except for the named entity extraction task which was not used.

\subsection{System performance}
\label{sec:sysPerf}

SLU systems can be evaluated with different metrics.
Historically, on the MEDIA corpus, two metrics are jointly used: the Concept Error Rate (CER) and the Concept/Value Error Rate (CVER).
The CER is computed similarly to the Word Error Rate, by only taking into account the concepts occurrences in both the reference and the hypothesis files.
The CVER metrics is an extension of the CER. It considers the correctness of the complete concept/value pair. In the example in section~\ref{sec:MEDIA}, both "one hundred and thirty" and \textsl{amount-payment} have to be correct to consider the concept/value pair (one hundred and thirty,\textsl{amount-payment}) as correct. Errors on the value component can come from a bad segmentation (missing or additional words in the value) or from ASR errors.

Table~\ref{tab:soa} presents the best results obtained on the official MEDIA benchmark dataset, by the main families of approaches presented in the two previous sections.
By computing the 95\% confidence interval, we observe a 0.7 confidence margin for CER and 0.8 for CVER, when the CER is 13.6\% and the CVER 18.5\%.
Until now, the best result was reached by an end-to-end encoder-decoder architecture with attention mechanism, trained by following a curriculum transfer-learning approach~\cite{9413581}.

\begin{table}[!htbp]
\centering
\begin{tabular}{|l|l|c|c|}
\hline
\textbf{Architecture}  &\textbf{Model}& \textbf{CER}& \textbf{CVER}\\
\hline
%\hline
%Manual & Camembert's based SLU~\cite{ghannay-2020-Coling}   & 7.56 & - \\
\hline
Cascade (2018)& HMM/DNN ASR + neural NLU \cite{simonnet-etal-2018-simulating} & 20.2 & 26.0\\
\hline
Cascade (2018) & HMM/DNN ASR + CRF \cite{simonnet-etal-2018-simulating} & 20.2 & 25.3\\
\hline
Cascade (2019) & HMM/TDNN ASR + CRF \cite{Caubriere2019} & 16.1 & 20.4 \\
\hline
End-to-end (2019) &E2E CTC ~\cite{Caubriere2019} & 16.4 & 20.9 \\
\hline
End-to-end  (2021)& E2E encoder-decoder with attention~\cite{9413581} & \textbf{13.6} & \textbf{18.5}\\

\hline
\end{tabular}
\caption{\label{tab:soa} Best results obtained on the official MEDIA benchmark dataset, by the main families of approaches presented in this paper.
Results are given in both Concept Error Rate and Concept/Value Error Rate} 
\end{table} 

%\ac{@Yannick : dans tableau 2 : est-ce qu'il ne faudrait pas ajouter les résultats cascade de notre hmm-tdnn + crf lors d'IS 19  (les res était de 16.1 CER + 20.4 CVER -- citation~\cite{Caubriere2019}}
%YE: oui, bien vu

% \subsection{Architectures}
%The proposed pipeline system is composed of two modules, the E2E ASR module to get automatic transcriptions, and the SLU module to annotate the automatic transcriptions with semantic tags. Those modules are described as follows.
% \subsection{E2E ASR}
%Fine tune wav2vec on commone Voice FR and ASR MEDIA .??

% For the E2E ASR, we investigate the use of models trained through self supervision in order to compute speech representations.
% We use the large French Wav2vec model \footnote{https://huggingface.co/LeBenchmark} \cite{evain2021lebenchmark}. 
% This model is pretrained using 3K hours of data. \\

% We propose to train our End to End speech recognition, by optimizing the learning rate for both the speech recognition and the self supervised representation. \\
% Our model is a neural architecture which gets as input a sequence of speech audio features and outputs character probabilities.
% The architecture used is a 

\section{Improving the state of the art}

In the previous section we present state-of-the-art performances.
Recently, unsupervised learning on huge amount of data have been successfully proposed to pre-train Transformers-based models~\cite{devlin-etal-2019-bert,baevski2020wav2vec}.
Thanks to these models, ASR state-of-the-art performance~\cite{baevski2020wav2vec} and NLP state-of-the-art performance~\cite{devlin-etal-2019-bert} have been outperformed, with respectively wav2vec and BERT models.
In this section, we present a cascade system using both BERT and wav2vec optimized on the MEDIA task. 

\subsection{BERT and CamemBERT models}

For the NLU module, we propose to use the one that achieved the state-of-the-art result on manual transcriptions of MEDIA corpus~\cite{ghannay-2020-Coling}.
This system is based on a fine-tuning of BERT~\cite{devlin-etal-2019-bert} on MEDIA SLU task using the French CamemBERT~\cite{martin2020CamemBERT} model.

BERT~\cite{devlin-etal-2019-bert} is a deeply bidirectional, unsupervised language representation model, which  stands  for Bidirectional Encoder Representations from Transformers.   
It is designed to pre-train deep bidirectional representations from unlabeled text, taking into account both left and right context in all layers.  
The resulting pre-trained BERT model can be fine-tuned with just one additional output layer, to create state-of-the-art models for a wide range of NLP tasks. 
BERT is pre-trained using a combination of masked language modeling objective and next sentence prediction on a large corpus which include the Toronto Book Corpus and Wikipedia.

The French CamemBERT model is based on RoBERTa (Robustly Optimized BERT Pre-training Approach)~\cite{liu2019roberta} which is based on BERT. 
CamemBERT is similar to RoBERTa, which dynamically change the masking pattern applied to the training data, and remove the next sentence prediction task. 
In addition, it uses the whole word masking and the SentencePiece tokenization \cite{kudo-richardson-2018-sentencepiece}. 
The CamemBERT model is trained on the French CCNet corpus composed of 135GB of raw text. 

\subsection{Wav2vec models}

Wav2vec 2.0 \cite{baevski2020wav2vec} is a model pre-trained through self-supervision. It takes raw audio as input and computes contextual representations that can be used as input for speech recognition systems. %\remsg{Tu peux dire contextual representations à la place de genral}
It contains three main components: a convolutional feature encoder, a context network and a quantization block.
The convolutional feature encoder converts the audio signal into a latent representation.
This representation is given to the context network which takes care of the context.
The context network architecture consists of a succession of several transformer encoder blocks. %\remsg{Est ce que le context network est composer de "several transformer " ? } \salima{oui}
The quantization network is used to map the latent representation to quantized representation.

%takes care of the context and also to a quantized to yield target context vectors.\\
%\salima{citer LeBenchmark}
%that is given to a context network which takes . \\
In \cite{evain2021lebenchmark}, the authors released French pre-trained wav2vec~2.0 models.
Two models have been released for public use\footnote{https://huggingface.co/LeBenchmark}, a large one and a base one.
In this study, we use the large configuration which encodes raw audio into frames of 1024-dimensional vectors. 
%The model is trained of about 300M parameters.
The models are pre-trained in a unsupervised way with 3K hours of unlabeled speech.
Details about data used to train the wav2vec models can be found in \cite{evain2021lebenchmark}.
The trained model is composed of about 300M parameters.

To get better ASR results than the ones we could reach by fine-tuning the French wav2vec~2.0 model, on the MEDIA training data only, we suggest to ,first, fine-tune on external audio data, as proposed in \cite{Caubriere2019} or \cite{9413581}.
To make the experiments reproducible, instead of using the Broadcast News data used in these works, we used the CommonVoice French dataset\footnote{https://commonvoice.mozilla.org/fr/datasets} (version 6.1), collected by the Mozilla Foundation, and much easily accessible.
The train set consists of 425.5 hours of speech, while the validation and test sets contain around 24 hours of speech.

\subsection{Cascade approach with pre-trained models}

As written before, we propose in this work to use a cascade approach, with pre-trained models for each component.
The ASR system is composed of the large pre-trained French wav2vec model, a linear layer of 1024 units, and the softmax output layer.
First, we optimize the ASR system on the French CommonVoice dataset.
Then, we fine-tune it for speech recognition on the French MEDIA corpus, the wav2vec weights being updated at each training stage. 
The loss function used at each fine-tuning step is the CTC loss function. 
We call the final ASR model W2V $\bullet $ Common Voice $\bullet$ $  M_{ASR}$.

%\remsg {il faut peut être dire que c'est basé sur 'l'approche CTC}

The NLU system is applied on the automatic transcriptions provided from the ASR system, to obtain semantic annotations.
This system is based on the fine-tuning of the French CamemBERT~\cite{martin2020CamemBERT} model, on the manual transcriptions of MEDIA corpus.
It achieved state-of-the-art result on manual transcriptions of MEDIA corpus~\cite{ghannay-2020-Coling}, yielding to 7.56 of CER when there is no error in the transcription.

\subsection{Results and discussion}

%In this study, the evaluation of the SLU slot filling task uses the Concept Error Rate (CER), which is estimated like the classical Word Error Rate (WER) but applied to semantic concepts instead of words. Insertions, substitutions, and deletions are all counted as errors. In addition to the Concept Value error rate (CVER), which is very close to CER but evaluates concept/value pairs instead of evaluating only concepts.

The experimental result obtained with the proposed cascade approach is presented in table~\ref{tab:proposedapproach}. 
We compare the performance of this cascade system, named \textit{W2V $\bullet $ Common Voice $\bullet$ $  M_{ASR}$ + CamemBERT}, to the E2E encoder-decoder model proposed in~\cite{9413581}, that reached the best result on this task until now, and other wav2vec-based models.
All the wav2vec-based models presented in table \ref{tab:proposedapproach} were implemented thanks to the SpeechBrain toolkit\footnote{https://speechbrain.github.io}, including the fine-tuning of the wav2vec models.

Like in section~\ref{sec:sysPerf}, the results are evaluated in terms of CER and CVER.
Our new system yields to 17.64\% of relative CER improvement and 7.02\% of relative CVER improvement, by reaching respectively 11.2\% of CER and 17.2\% of WER.
The result shows the effectiveness of unsupervised pre-trained models like wav2vec and BERT in such a scenario.
Notice that the \textit{W2V $\bullet $ Common Voice $\bullet$ $  M_{ASR}$} model allows us to have an effective ASR system that achieved 8.5\% of WER. 

In system (1), the wav2vec model is fine-tuned directly on MEDIA SLU ($M_{SLU}$) task.
In system (2), the wav2vec model is first fine-tuned on the Common Voice data then on $M_{SLU}$ task, and a beam search decoding is applied.
In systems (3) and (4) the wav2vec model is first fine-tuned on the Common Voice data, then on MEDIA ASR ($M_{ASR}$), and last on $M_{SLU}$ task, using the greedy or the beam search decoding using a 5-gram language model to rescore.  
This language model is trained on the manual transcriptions of $M_{SLU}$ training data only.

It is worth to mention that even before the generalisation of the use of neural networks for sequential tagging tasks, such as the slot filling task investigated in this paper, several efforts have been made to better take into account the  the ASR system errors during the semantic labeling.
Many approaches have been proposed for a joint decoding between speech recognition and understanding, considering the n-best recognition hypotheses during the semantic annotation \cite{hakkani2005,servan2006,jabaian2013}.
When neural networks have become state-of-the-art systems for SLU, end-to-end approaches have gradually replaced cascade approaches and have shown very good performance, allowing the semantic labeling of a speech signal and minimising the impact of transcription errors on the SLU performance. 
However, these architectures need a large amount of data and often use pre-trained external module that have been trained separately in out-of-context data. The results presented in table \ref{tab:proposedapproach} show that if such pre-trained models are used in a cascade architecture, the resulting system reaches or even exceeds the performance of the end-to-end based one. 
%\remsg{J'ai ajouté cette phrase, qu'en pensez vous?}
In addition, the result of the cascade system reinforces the idea of the use of pre-trained models at the encoder (wav2vec) and decoder (BERT) levels within end-to-end architecture, as proposed in~\cite{chung2021splat}. This leads us to conclude that the two architectures remain valid and competitive and that the choice should be made according to the availability of additional data and the pre-training models.

\begin{table}[!htb]
\centering
\begin{tabular}{|l|l|c|c|}
\hline
\textbf{Architecture}  &\textbf{Model}  & \textbf{CER}& \textbf{CVER}\\
\hline
%\hline
%Camembert's based SLU~\cite{ghannay-2020-Coling}   &7.56 & - \\
\hline
\multirow{5}{*}{End-to-end} & encoder-decoder~\cite{9413581} & 13.6 & 18.5\\
\cline{2-4}
%\hline
%ASR $\bullet$ MEDIA SLU (BEAM) ~\cite{Caubriere2019} & 20.1 & 24.0 \\
%\hline
%marco & &\\
%\hline
%E2E SLU ED/A & 47.4  & 69.1  \\small architecture
%E2E SLU ED/A & 57.8  & 78.3  \\ 
%\hline
%E2E SLU pure CTC / greedy mode &   &   \\
%\hline
%E2E SLU pure CTC / % beamsearch mode &   &   \\
%\hline
%ASR common Voice Pretrain $\bullet$ E2E SLU &  35.9& 57.6\\
%\hline
 %\hline
 
 %% AC TODO : check YE -  finetuning sur MEDIA SLU du modèle : /local_disk/calypso/yesteve/SLU/exp/w2v/results/debug/simpleNN_ASR_wav2vec_ctc/1242/save/wav2vec2_checkpoint/model.pt
& (1) W2V $\bullet$ $M_{SLU}$ (Beam 5g) & 18.8 & 23.6 \\
& (2) W2V $\bullet $ Common Voice $\bullet$ $M_{SLU}$ (Beam 5g)& 15.8 & 20.4 \\

& (3) W2V $\bullet $ Common Voice $\bullet$ $M_{ASR}$ $\bullet$ $M_{SLU}$ (greedy) & 15.4 & 20.5 \\

& (4) W2V $\bullet $ Common Voice $\bullet$ $  M_{ASR}$ $\bullet$ $M_{SLU}$ (Beam 5g) & 14.5 & 18.8 \\
\hline
\hline
Cascade & W2V $\bullet $ Common Voice $\bullet$ $  M_{ASR}$ + CamemBERT & \bf 11.2 & \bf 17.2\\
\hline
\end{tabular}
\caption{\label{tab:proposedapproach} Performance on Test MEDIA in terms of CER and CVER scores of the proposed cascade and end-to-end systems using pre-trained models. "$\bullet$" formalizes a transfer learning step during the training of the E2E system.} 
\end{table}

\section{Conclusions}
In this paper, we present a brief overview of the recent advances on the French MEDIA benchmark dataset for SLU. 
We propose a system based on a cascade approach, that takes benefit from acoustic-based and linguistic-based models pre-trained on unlabelled data : wav2vec models for the ASR system, and BERT-like model for the NLU system. 
Experimental results show that our system  outperforms significantly the current state of the art with a Concept Error Rate (CER) of 11.2\% instead of 13.6\% for the last state-of-the-art system presented this year.

%More, we observe that when the pre-trained models are used in a cascade architecture, the resulting system exceeds the performance of the end-to-end based one.
%\ac{"the performance of the current end-to-end SoA" ? la phrase donne l'impression qu'on a fait un end-to-end a base de wav2vec / bert (même si la phrase d'après enlève le doute)}

This new advance reinforces the idea of the use of pre-trained models at the encoder (wav2vec) and decoder (BERT) levels within an end-to-end architecture. This will be explored in our future work.

This study leads us to conclude that the two architectures (cascade \textsl{vs.} end-to-end) remain valid and competitive and that the choice should be made according to the availability of additional data and relevant pre-trained models.

\bibliographystyle{splncs04}
 \bibliography{mybib}

\end{document}